\documentclass[runningheads]{llncs}
\usepackage{amsmath,graphicx}

\usepackage{enumitem}
\setlist{nosep, leftmargin=14pt}
\DeclareMathAlphabet\mathbfcal{OMS}{cmsy}{b}{n}

\usepackage{subcaption}
\usepackage{graphicx}
\usepackage{hyperref}
\usepackage{multirow}
\usepackage{tabularx}
\newcolumntype{Y}{>{\centering\arraybackslash}X}
\newcolumntype{j}{X}
\newcolumntype{s}{>{\hsize=.35\hsize}Y}
\newcolumntype{?}{!{\vrule width 1pt}}

\usepackage{mwe} 

\begin{document}
%
\title{The Effect of the Loss on Generalization: Empirical Study on Synthetic Lung Nodule Data}

\titlerunning{The Effect of the Loss on Generalization}

%
\author{Vasileios Baltatzis \inst{1, 2}, Lo\"ic Le Folgoc \inst{2},  Sam Ellis \inst{1},  Octavio E. Martinez Manzanera \inst{1}, Kyriaki-Margarita Bintsi \inst{2}, Arjun Nair \inst{3}, Sujal Desai \inst{4}, \\ Ben Glocker \inst{2}, Julia A. Schnabel \inst{1, 5, 6}}

\authorrunning{V. Baltatzis et al.}
%
\institute{School of Biomedical Engineering and Imaging Sciences, King’s College London, UK \\
\and BioMedIA, Department of Computing, Imperial College London, UK \\
\and Department of Radiology, University College London, UK \\
\and The Royal Brompton \& Harefield NHS Foundation Trust, London, UK \\
\and Technical University of Munich, Germany \\
\and Helmholtz Center Munich, Germany\\
\email{vasileios.baltatzis@kcl.ac.uk}
}
\maketitle              
\begin{abstract}
Convolutional Neural Networks (CNNs) are widely used for image classification in a variety of fields, including medical imaging. While most studies deploy cross-entropy as the loss function in such tasks, a growing number of approaches have turned to a family of contrastive learning-based losses. Even though performance metrics such as accuracy, sensitivity and specificity are regularly used for the evaluation of CNN classifiers, the features that these classifiers actually learn are rarely identified and their effect on the classification performance on out-of-distribution test samples is insufficiently explored. In this paper, motivated by the real-world task of lung nodule classification, we investigate the features that a CNN learns when trained and tested on different distributions of a synthetic dataset with controlled modes of variation. We show that different loss functions lead to different features being learned and consequently affect the generalization ability of the classifier on unseen data. This study provides some important insights into the design of deep learning solutions for medical imaging tasks.
\end{abstract}
\section{Introduction}
\label{sec:intro}

\begin{figure}[t!]
\centering
\includegraphics[scale = 0.15]{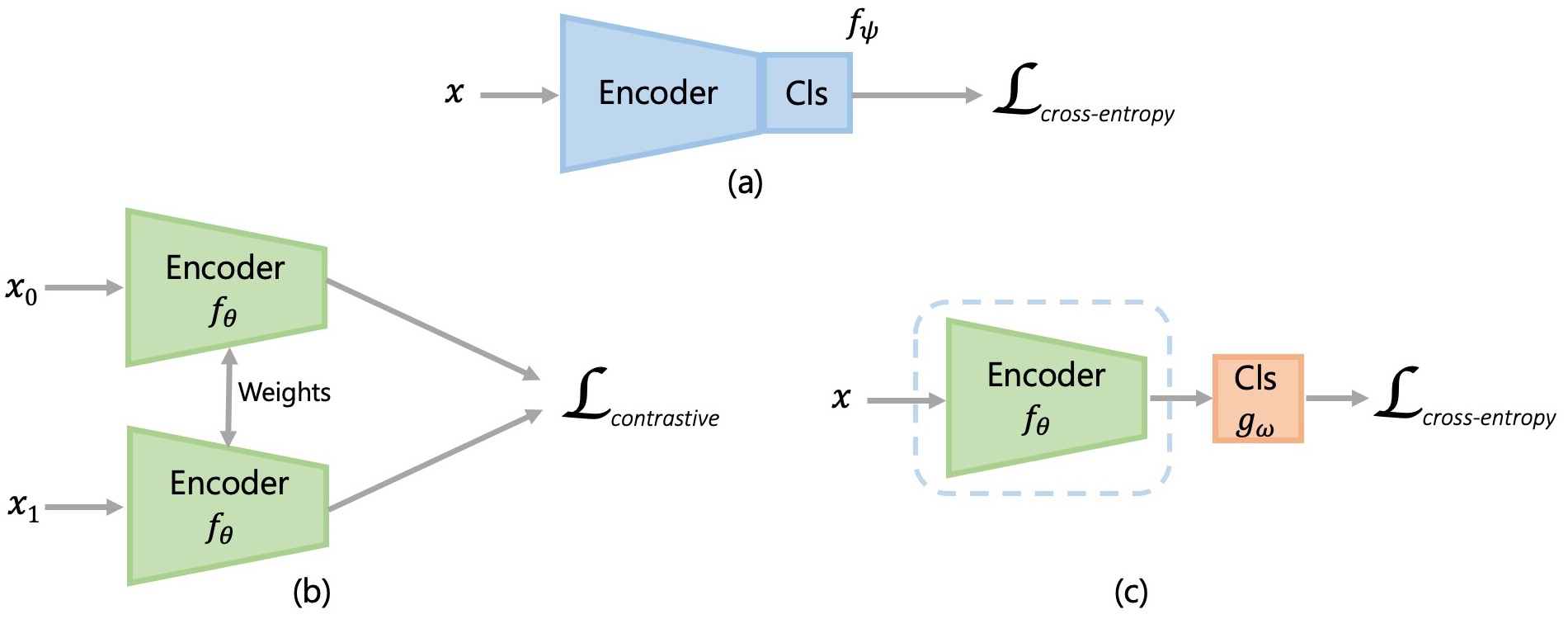}
\caption{Overview of the used CNN architectures, which are based on LeNet-5 \cite{LeCun1998Gradient-basedRecognition}. (a) An encoder-classifier network trained end-to-end with CE. (b) A siamese network trained with a contrastive loss. (c) The encoder from (b) is frozen and a classifier (Cls), identical to the one used in (a), is trained on top of it with CE. } 
\label{networks}
\end{figure}

Deep learning methods and particularly Convolutional Neural Networks (CNNs) are, currently, the backbone of most state-of-the-art approaches for medical image classification tasks. The performance of machine learning techniques, however, can drop significantly when the test data are from a different distribution than the training data, which is common in real-world applications, such as medical images originating from different hospitals, acquired with different protocols, or when there is a lack of or variation in high quality annotations. 

Motivated by these obstacles, we study the effect of data variation utilizing a synthetic dataset with specific modes of variation.The limitations that arise from a synthetic dataset are clear since its simplified nature does not reflect the complexity of a real, clinical dataset. However, it is exactly this complexity that we are trying to avoid, as it would not allow us to evaluate very specific scenarios in terms of controlling the exact characteristics of the training and test distributions. This fully controlled setting allows us to create training and test distributions with similar or contrasting characteristics. We leverage this dataset to explore the subtlety of the differences between training and test distributions that is sufficient to hamper performance. We do not suggest that this simplification can lead to a direct application on disease classification but rather our primary intent is to investigate the behavior of CNNs under certain distribution shifts at test time to a very fine level of detail, which would be impossible to achieve if we shifted to a real-world medical imaging dataset. To examine thoroughly the features learned by a CNN and how these can influence the performance for out-of-distribution (OOD) test samples, we utilize principal component analysis (PCA) and saliency maps. Additionally, we study the increasingly popular contrastive learning-based losses \cite{Hadsell2006DimensionalityMapping} proposed in recent work \cite{Dou2019DomainFeatures,Winkens2020ContrastiveDetection}. Here, we investigate the differences between a cross-entropy (CE) loss and a contrastive loss, in terms of both performance and resulting CNN features.

Our contributions can be summarized as follows:
1) We design a synthetic dataset with two modes of variation (binary shape class and average intensity of the shape appearance) inspired by the real world application of lung nodule classification;
2) We conduct an experimental study to explore the effects of two different loss functions (CE and contrastive) on the learned CNN features (under different training distributions) and the impact on OOD generalization;
3) We use a variety of performance metrics (accuracy, sensitivity, specificity) and visualizations (PCA, saliency maps) to support and evaluate our findings. Our findings and insights will be of interest to practitioners designing machine learning solutions for medical imaging applications.

\section{Materials and Methods}
\label{sec:format}


\subsection{Data}
\label{ssec:data}
The synthetic dataset used here is inspired by the real world application of lung nodule classification and is designed based on two modes of variation. The first mode is the shape class, which is binary. Abnormalities, such as spikes, on the perimeter of lung nodules are termed as spiculation and often indicate malignancy, while a smoother outline is often associated with benign disease \cite{McWilliams2013ProbabilityCT}. We refer to the two classes as malignant and benign, to form a paradigm similar to lung nodules. The second mode of variation is appearance represented by the average intensity of the pixels within each shape. The values range from 110 to 200 with noise added in 10-point increments, thus giving 10 possible values for this mode, while the background intensities remain fixed for all samples. The synthetic data have been constructed by manually drawing two base shapes (benign vs malignant) from which the experimental dataset is generated using random spatial transformations produced by a combined affine and non-rigid FFD-based transformation model. With $\mathcal{D}_{mode}(i_{mal}, i_{ben})$, we denote a distribution where the average foreground intensity of the malignant and benign shapes is $i_{mal}$ and $i_{ben}$ respectively, while $mode$ refers to either the training or the test set. 

\subsection{Neural network architectures and loss functions}

We consider two different losses, a CE loss and a contrastive loss, and consequently two neural network architectures that facilitate the two losses (Figure \ref{networks}). For simplicity we consider a binary classification task. Both architectures are based on the well-established LeNet-5 \cite{LeCun1998Gradient-basedRecognition}. For the first approach, we use a combined encoder-classifier network $f_{\psi}$ with parameters $\psi$. It is trained end-to-end, given input image $X$ and label $y$, via the CE loss (Eq. \eqref{crossentropy}):

\begin{equation}
\mathcal{L}_{CE} = -y \text{ } log(f_{\psi}(X)) - (1-y)log(1-f_{\psi}(X))
\label{crossentropy}
\end{equation}

For the second approach we use a Siamese network as in \cite{Hadsell2006DimensionalityMapping}, trained in two stages. In the first stage, the network is composed of two copies of the encoder $f_{\theta}$ that share the same weights $\theta$. The input for this system is a pair of images $(X_0,X_1)$ with labels $(y_0, y_1)$ that go through the encoders to produce the representations $f_{\theta}(X_0)$ and $f_{\theta}(X_1)$, which are then fed into the contrastive loss defined in Eq.~\eqref{contrastive}:

\begin{equation}
\mathcal{L}_{contr} =  \\
\begin{cases}
    d_\theta(X_0, X_1)^2, &\text{if }y_0 = y_1  \\
     \{max(0, m-d_\theta(X_0, X_1)\}^2, &\text{if }y_0 \neq y_1 \\
\end{cases}
\label{contrastive}
\end{equation}

\begin{equation}
\text{where } d_\theta(X_0, X_1) = \lVert f_{\theta}(X_0)-f_{\theta}(X_1) \rVert_2\, .
\label{distance}
\end{equation}

The loss function minimizes the representation-space distance of Eq.~\eqref{distance} between samples of the same class, while maximizing (bounded by the margin $m$) the distance between samples of different classes.
In the second stage, the encoder $f_{\theta}$ is frozen. We then add a classifier $g_{\omega}$, with parameters $\omega$, that uses the representations $f_{\theta}(X)$ as input to perform the classification task. Similarly to the first approach, the encoder $f_{\theta}$ uses an image $X$ and a label $y$ as input and the classifier $g_{\omega}$ is trained with the CE loss. This way, the contrastive loss is used to pre-train the encoder of the network, thus leading to a different set of features that is used for the classification task, compared to the first approach where training is end-to-end.


\section{Experiments and Results}
\label{sec:experiments-results}


\begin{table}[t]
\caption{Quantitative results for the three experimental scenarios. We report accuracy (Acc), sensitivity (SE) and specificity (SP), for each of the training distribution ($\mathcal{D}_{tr}$), test distribution ($\mathcal{D}_{te}$) and loss combinations that are described in section \ref{sec:experiments-results}. The results that appear on the table correspond to the performance on ($\mathcal{D}_{te}$). For the training set, all metrics have a value of 1.00 and therefore are not reported on the table.}
\label{table:1}
\centering
\begin{tabularx}{\textwidth}{|j|sss||sss|}
\hline
{}&$\mathbfcal{D}_{\textbf{tr}}$& $\mathbfcal{D}_{\textbf{te}}$ & \textbf{Loss} & \textbf{Acc} & \textbf{SE} & \textbf{SP} \\ 
\hline
\multirow{4}{*}{Experimental Scenario 1}&150,150 & 130,170  &CE& 1.00& 1.00 & 1.00 \\
{}&150,150 & 170,130  &CE& 0.62& 0.87 & 0.37 \\
{}&150,150 & 130,170 & Contrast & 1.00 & 1.00 & 1.00 \\
{}&150,150 & 170,130 & Contrast & 0.15 & 0.30 & 0.00 \\
\hline
\multirow{4}{*}{Experimental Scenario 2}&180,160 & 150,190  &CE& 0.94& 0.90 & 0.98 \\
{}&180,160 & 190,150  &CE& 0.96& 0.94 & 0.98 \\
{}&180,160 & 150,190 & Contrast & 0.27 & 0.01 & 0.53 \\
{}&180,160 & 190,150 & Contrast & 1.00 & 1.00 & 1.00 \\
\hline 
Experimental Scenario 3&180,150 & 150,190 & CE& 0.59 & 0.35 & 0.83 \\ 
\hline
\end{tabularx}
\end{table}

\begin{figure*}[t!]
\centering
  \begin{subfigure}[t]{0.45\linewidth}
    \centering
    \includegraphics[width =\textwidth]{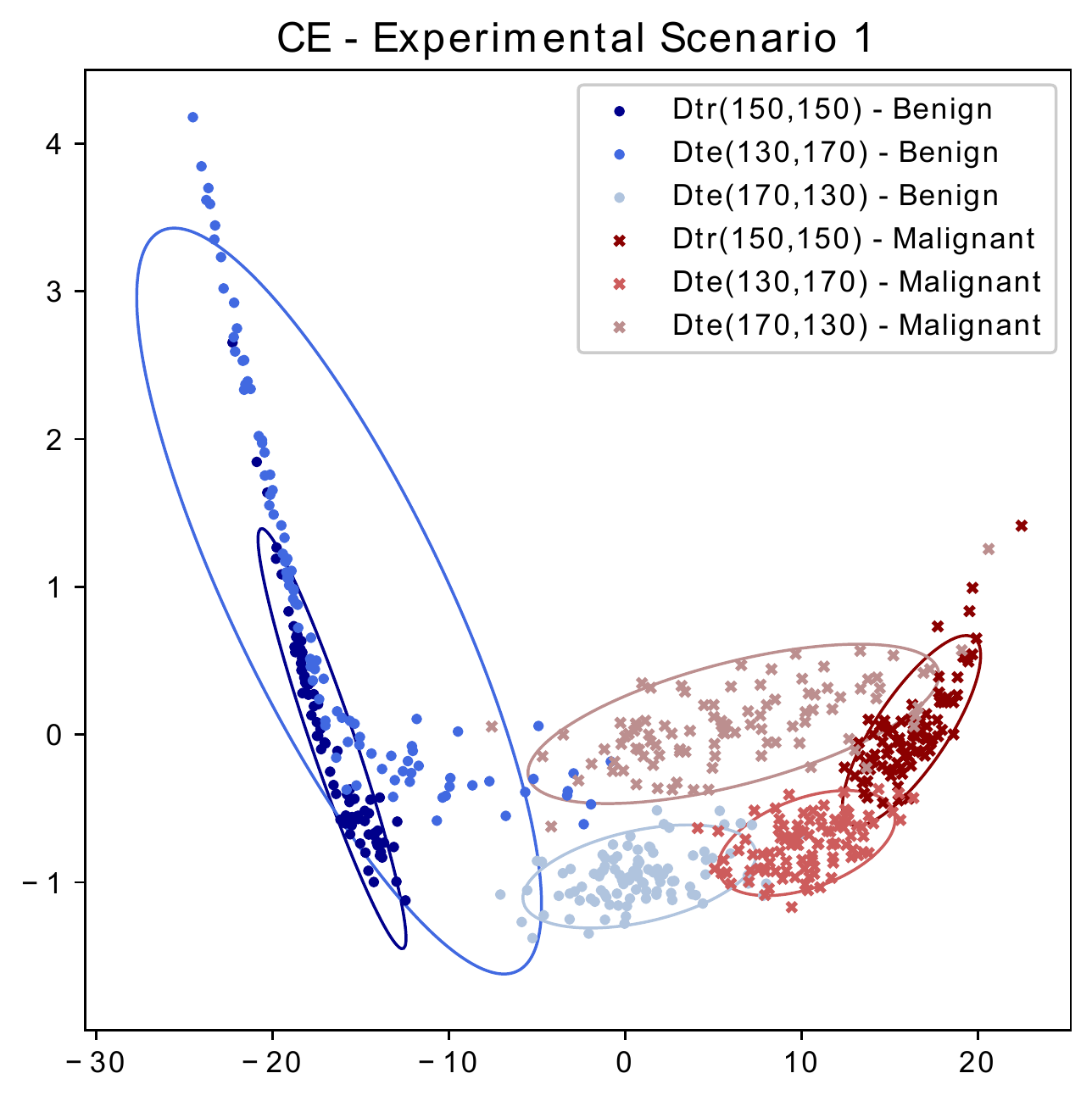}\hfill%
    \caption{} 
    \label{pca:a} 
  \end{subfigure}
  \begin{subfigure}[t]{0.48\linewidth}
    \centering
    \includegraphics[width =\textwidth]{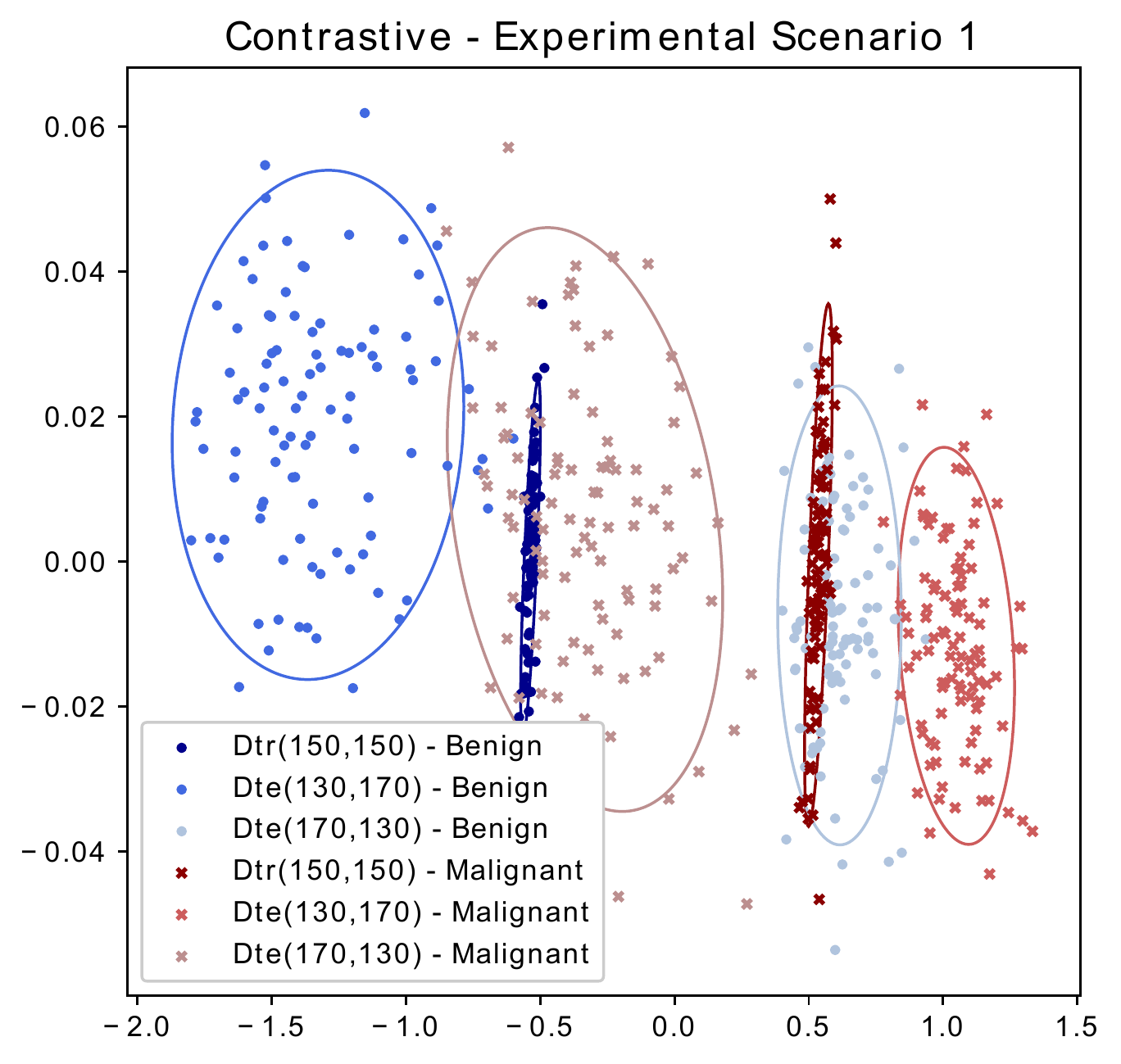} 
    \caption{} 
    \label{pca:b} 
  \end{subfigure} 
  \begin{subfigure}[t]{0.45\linewidth}
    \centering
    \includegraphics[width =\textwidth]{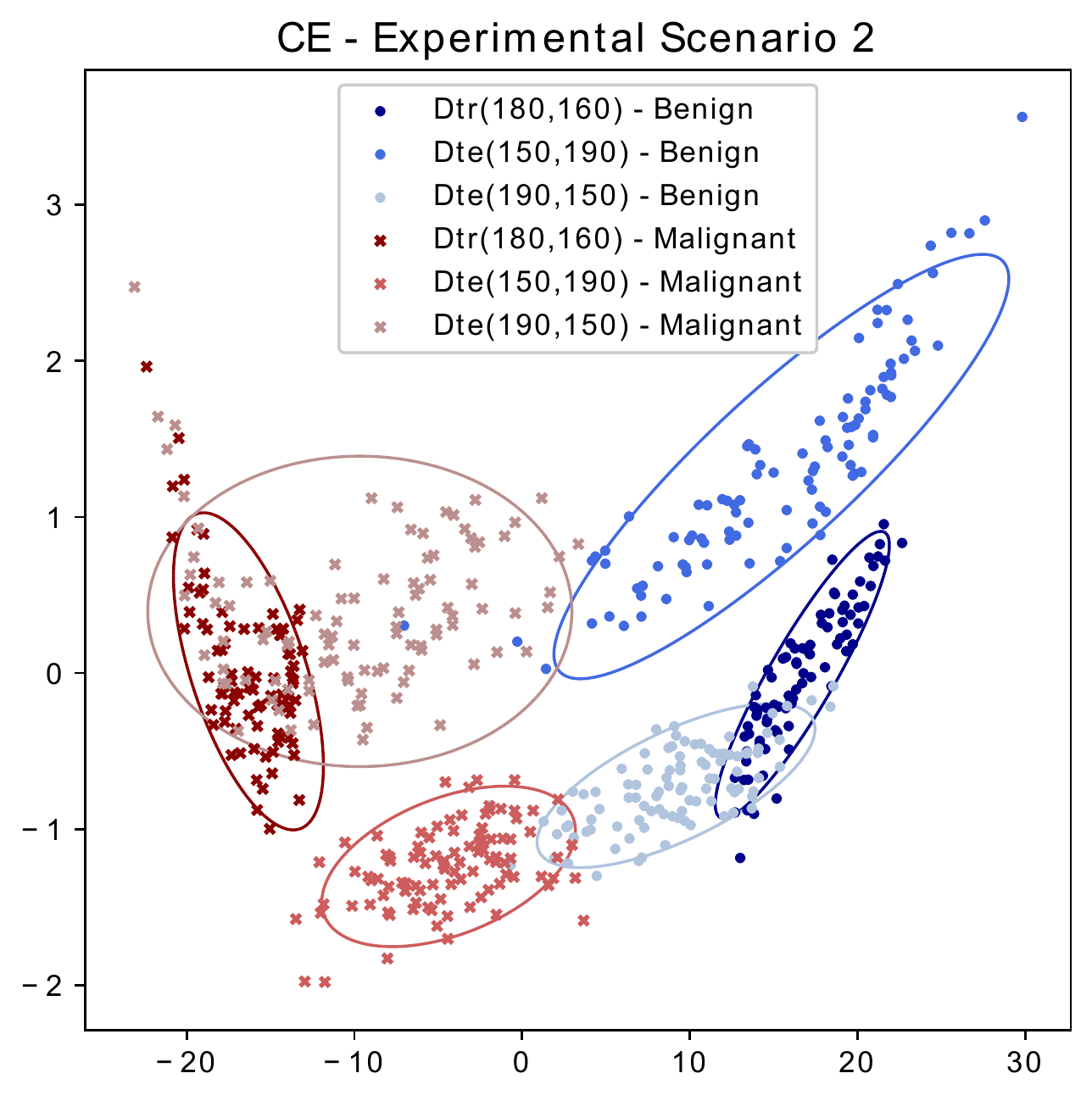}\hfill%
    \caption{} 
    \label{pca:c} 
  \end{subfigure}
  \begin{subfigure}[t]{0.485\linewidth}
    \centering
    \includegraphics[width =\textwidth]{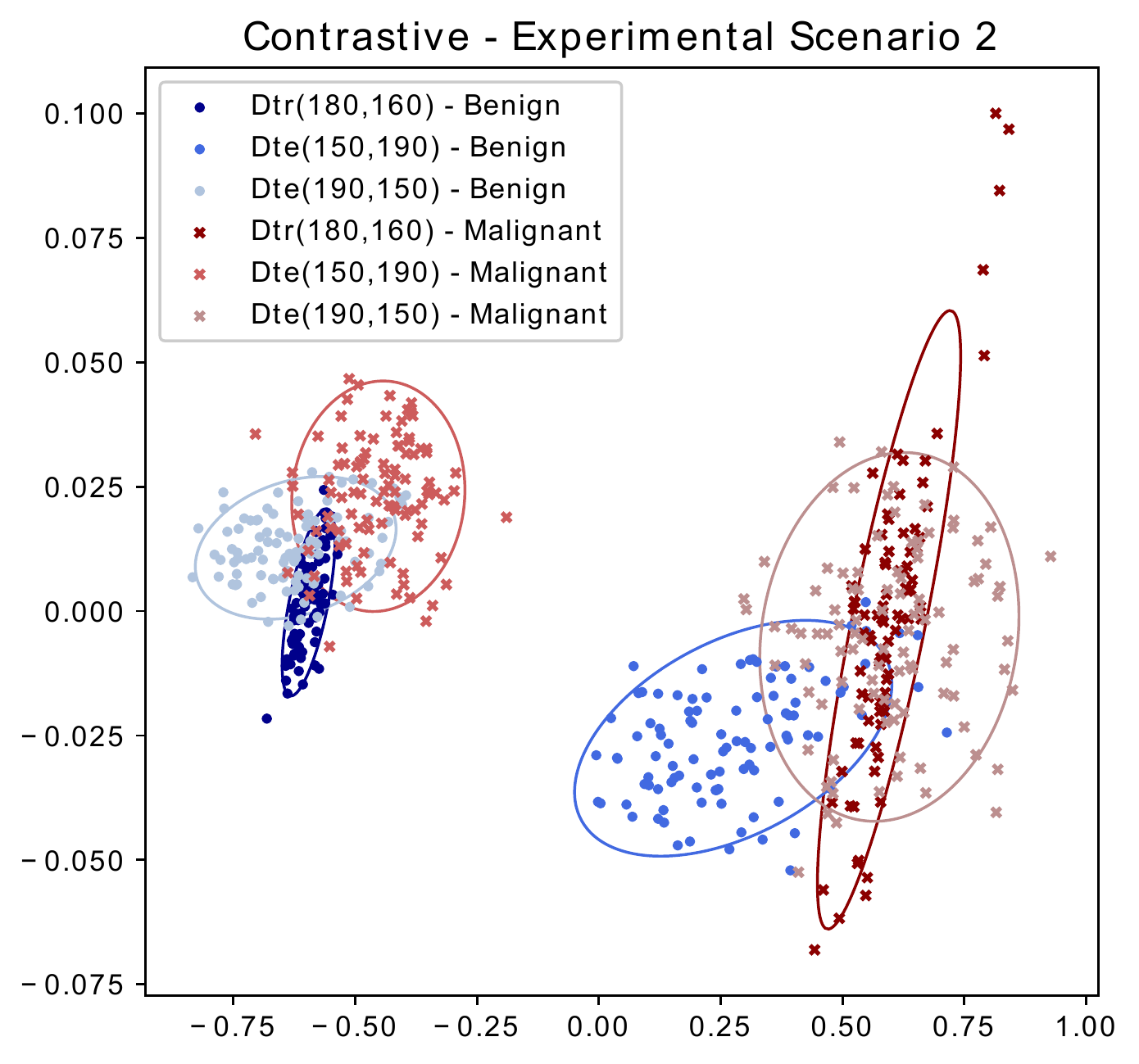}
    \caption{} 
    \label{pca:d} 
  \end{subfigure} 
  \caption{Experimental Scenario 1 (CE (a), Contrastive (b)) and 2 (CE (c), Contrastive (d)): The PCA projections of the last layer of the CNN before the classification layer. Red hues/'x' denote malignant samples while blue hues/'o' benign. Dark hues are used for the training set $\mathcal{D}_{tr}(i_{mal}, i_{ben})$ and light hues for the two test sets $\mathcal{D}_{te}(i_{mal}, i_{ben})$. The ellipsoids mark two standard deviations distance from the mean of each distribution. In (a) the benign samples of $\mathcal{D}_{te}(170, 130)$ are closer to the malignant cluster of $\mathcal{D}_{tr}(150, 150)$ leading to low specificity. The same is happening in (b) but the malignant samples are also closer to the benign cluster and hence overall performance is low. In (c) there is good generalization in both OOD test sets. In (d) both benign and malignant samples of $\mathcal{D}_{te}(150, 190)$ are close to the opposite cluster of $\mathcal{D}_{tr}(180, 160)$ leading to low performance.}
  \label{pca_exp12} 
\end{figure*}

We devise three experimental scenarios to demonstrate the OOD test performance by controlling different aspects of the training distribution. For quantitative evaluation, we use accuracy, sensitivity and specificity. We only report these metrics for the OOD test sets, since at train time they are all 1.00. For qualitative evaluation, we utilize PCA to get a two-dimensional projection of the last layer of the CNN before the classification layer and explore the learned feature space. We also use gradient saliency maps \cite{Simonyan2014DeepMaps} to investigate the areas of the input image that contribute most to the CNN prediction. 

\textbf{Training details} We draw 200 samples from the training distribution, $85\%$ of which are for training and $15\%$ for validation, and another 200 samples from the test distribution for testing. The networks are trained using the Adam optimizer \cite{Kingma2015Adam:Optimization} ($\text{learning rate} =10^{-4}$) for 100 epochs and a batch size of 32 samples. The positive and negative pairs for the contrastive loss are dynamically formed within each batch. The margin is chosen to be $m=1$ based on validation performance, and the Euclidean distance is used as the distance metric. All experiments were conducted using PyTorch \cite{Paszke2019PyTorch:Library} and the models were trained on a Titan Xp GPU.

\subsection*{Experimental Scenario 1} 
\begin{figure*}[t!]
\centering
  \begin{subfigure}[t]{0.40\linewidth}
    \centering
    \includegraphics[width =\textwidth]{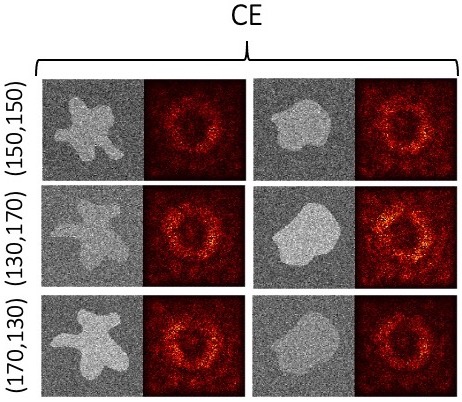}
    \caption{} 
    \label{saliency:a} 
  \end{subfigure}
  \begin{subfigure}[t]{0.37\linewidth}
    \centering
    \includegraphics[width =\textwidth]{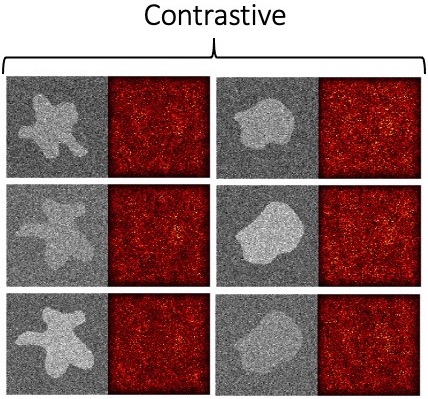} 
    \caption{} 
    \label{saliency:b} 
  \end{subfigure} 
  \begin{subfigure}[t]{0.40\linewidth}
    \centering
    \includegraphics[width =\textwidth]{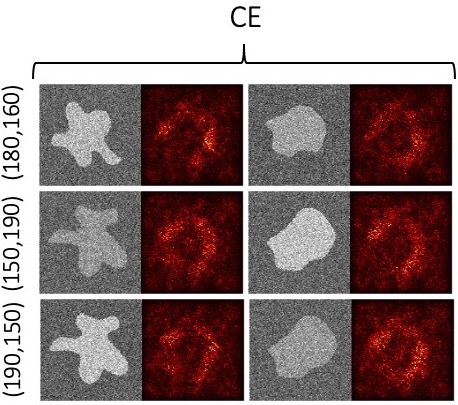}
    \caption{} 
    \label{saliency:c} 
  \end{subfigure}
  \begin{subfigure}[t]{0.38\linewidth}
    \centering
    \includegraphics[width =\textwidth]{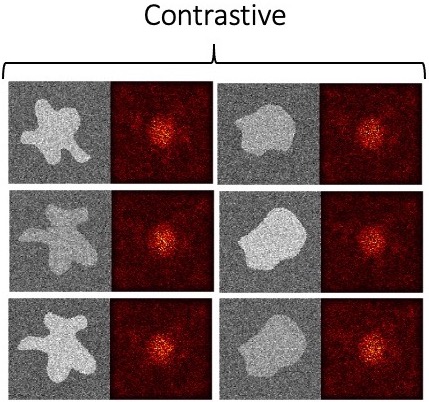}
    \caption{} 
    \label{saliency:d} 
  \end{subfigure} 
  \caption{Experimental Scenario 1 (CE (a), Contrastive (b)) and 2 (CE (c), Contrastive (d)): Example images along with their saliency maps. The first row corresponds to the training set, while the other two to the OOD test sets. In (a) and (b) the activations are spread out across the whole image, even though in (a) some patterns regarding the shape are being picked up. In (c) the activations are strong around the border of the shape, while the background activations are a bit lower compared to (a). In (d) the activations are at their highest at the centre of the image (i.e. within the shape).}
  \label{saliency12} 
\end{figure*}

Initially, we consider the case where malignant and benign shapes have the same average intensity ($i_{mal} = i_{ben}$). Specifically, we select $\mathcal{D}_{tr}(150, 150)$, since 150 is an intensity in the middle of the distribution of the available intensities, and we use $\mathcal{D}_{te}(130, 170)$ and $\mathcal{D}_{te}(170, 130)$ as these intensities have equal distance from the training distribution for both malignant and benign shapes. Performance metrics can be found in the top four rows of Table \ref{table:1}; PCA projections and saliency maps in Figures \ref{pca:a}, \ref{pca:b} and \ref{saliency:a}, \ref{saliency:b}, respectively. The CNN fails to classify the OOD test correctly when $i_{mal} > i_{ben}$ for either loss.


\subsection*{Experimental Scenario 2}
Next, we consider the case where the average intensities of the whole image (i.e. including the background and not just the pixels inside the shape) are equal for benign and malignant samples ($i_{global\_mal} = i_{global\_ben}$). This happens for $\mathcal{D}_{tr}(180, 160)$, where the average whole image intensity for both malignant and benign images is 117. Equivalently to the first scenario, the OOD test sets come from $\mathcal{D}_{te}(150, 190)$ and $\mathcal{D}_{te}(190, 150)$. The CE trained CNN is able to generalize on both OOD test datasets, while the contrastive loss trained CNN fails when the relationship between $i_{mal}$ and $i_{ben}$ is opposite to what it was in the training distribution. The quantitative results are reported in rows 5-8 of Table \ref{table:1}, while the qualitative results are visualized in Figures \ref{pca:c}, \ref{pca:d} (PCA) and \ref{saliency:c}, \ref{saliency:d} (saliency).


\subsection*{Experimental Scenario 3} 
With the final experiment we want to focus just on one single finding which is the effect of the smallest possible change to the training distribution of the previous scenario (i.e. $\mathcal{D}_{tr}(180, 150)$ instead of $\mathcal{D}_{tr}(180, 160)$), while retaining the same test distributions. For simplicity, we do not focus on analyzing the behaviour of the CNN feature space through saliency maps and PCA projections nor do we use the contrastive loss. We just show results for the CE loss to make sure that we highlight the drop in performance from 0.94 to 0.59 (last row of Table \ref{table:1}) even with the smallest of changes.


\section{Discussion}
\label{sec:discussion}
There are three underlying features in the synthetic data distribution that a CNN can try to capture. These are the average intensity of the whole image, the average intensity of the foreground pixels and the shape of the object. From the results of Experimental Scenario 1, where the foreground intensities are equal at train time, we observe that for both losses the CNN fails when the malignant intensity is higher than the benign intensity at test time. This is happening because in this setting, the whole image average intensity is lower for malignant (110) than benign (114) samples, due to the more convex shape of the benign samples, which allows for fewer background pixels. Consequently, the CNN can easily pick up on that feature to distinguish the two classes regardless of the loss function. This can be also confirmed by the saliency maps (Figures \ref{saliency:a},\ref{saliency:b}), where the activations are spread throughout the whole image, especially for the contrastive loss. The CE loss appears to pick up some patterns in the border of the shape, but the separation of the PCA projections between the two classes is no longer clear for $\mathcal{D}_{te}(170, 130)$ (Figure \ref{pca:a}).

In Experimental Scenario 2, we remove this discrepancy in the global intensities, and therefore the CNN can no longer use that as a discriminatory feature. In that case, the CNN that was trained with CE is able to generalize in both OOD test sets, which can be confirmed by the PCA projections as well, as they retain the same spatial location as in the training set (Figure \ref{pca:c}). Hence, it must be capturing the shape information itself. On the other hand, the CNN trained with the contrastive loss learns to distinguish samples based on the average intensity of the pixels of the shape itself, which is evident from the saliency maps, where the most important pixels are the ones in the center of the image (i.e. within the shape) (Figure \ref{saliency:d}). Therefore, the CNN fails when $i_{mal} < i_{ben}$ at test time, since it was $i_{mal} > i_{ben}$ at train time, and the PCA projections for $\mathcal{D}_{te}(150, 190)$) have the opposite mapping to the one for either $\mathcal{D}_{tr}(180, 160)$) or $\mathcal{D}_{te}(190, 150)$).

Finally, in Experimental Scenario 3, we demonstrate that even the slightest change (i.e. reduce $i_{ben}$ to 150 from 160) can have a dramatic impact on the performance of the model on OOD test data, as the accuracy drops from 0.93 to 0.59 for $\mathcal{D}_{te}(150, 190)$. These results indicate how unreliable CNNs can be even when tested on data that are not that far from the training distribution. We demonstrate this failure on a relatively simple dataset. In real applications the relationship between features and the task at hand can be expected to be more complex leading to even worse OOD generalization.

\section{Conclusion}
\label{sec:conclusion}
Motivated by the important clinical application of lung nodule classification, we have designed a synthetic dataset from a controlled set of variation modes and conducted an exploratory analysis to obtain insights into the learned feature space when trained on different parts of the dataset distribution and how this affects the OOD generalization. The findings indicate that CNN predictions are initially based on the whole image average intensity. When this effect is prohibited, the CNN trained with CE focuses on shape, while the contrastive loss leads the CNN to pick up the average intensity of foreground pixels. Moving forward, we will explore how to constrain the feature space in an automated manner by incorporating application-specific prior knowledge and apply this approach on clinical data.





\section{Acknowledgments}
\label{sec:acknowledgments}
This work is funded by the King’s College London \& Imperial College London EPSRC Centre for Doctoral Training in Medical Imaging (EP/L015226/1), EPSRC grant EP/023509/1, the Wellcome/EPSRC Centre for Medical Engineering (WT 203148/Z/16/Z), and the UKRI London Medical  Imaging \& Artificial Intelligence Centre for Value Based Healthcare. The Titan Xp GPU was donated by the NVIDIA Corporation. 

\bibliographystyle{splncs04}
\bibliography{main}

\end{document}